# IGAN: A New Inception-based Model for Stable and High-Fidelity Image Synthesis Using Generative Adversarial Networks

Ahmed A. Hashim [1], Ali Al-Shuwaili[2], Asraa Saeed[3], Ali Al-Bayaty[4*]

[1] College of Business Informatics, University of Information Technology and Communications, Baghdad, Iraq, dr.ahmed.hashim@uoitc.edu.iq
[2] College of Engineering, University of Information Technology and Communications, Baghdad, Iraq, ali.najdi@uoitc.edu.iq
[3] Information Institute for Postgraduate Studies, University of Information Technology and Communications, Baghdad, Iraq, phd202110680@iips.edu.iq
[4] Department of Electrical and Computer Engineering, Portland State University, Portland, OR, USA, albayaty@pdx.edu

**Abstract**

Generative Adversarial Networks (GANs) face a significant challenge of striking an optimal balance between high-quality image generation and training stability. Recent techniques, such as DCGAN, BigGAN, and StyleGAN, improve visual fidelity; however, such techniques usually struggle with mode collapse and unstable gradients at high network depth. This paper proposes a novel GAN structural model that incorporates deeper inception-inspired convolution and dilated convolution. This novel model is termed the Inception Generative Adversarial Network (IGAN). The IGAN model generates high-quality synthetic images while maintaining training stability, by reducing mode collapse as well as preventing vanishing and exploding gradients. Our proposed IGAN model achieves the Fréchet Inception Distance (FID) of 13.12 and 15.08 on the CUB-200 and ImageNet datasets, respectively, representing a 28–33% improvement in FID over the state-of-the-art GANs. Additionally, the IGAN model attains an Inception Score (IS) of 9.27 and 68.25, reflecting improved image diversity and generation quality. Finally, the two techniques of dropout and spectral normalization are utilized in both the generator and discriminator structures to further mitigate gradient explosion and overfitting. These findings confirm that the IGAN model potentially balances training stability with image generation quality, constituting a scalable and computationally efficient framework for high-fidelity image synthesis.

## 1. Introduction

One of the major challenges in Generative Adversarial Network (GAN) research is to simultaneously improve the overall training stability and generative capability [1-2]. Simply adjusting the overall number of model parameters across different scenarios is not a feasible solution. While increasing the number of such parameters can enhance the model's features extraction ability and improve the quality of generated images, as seen in models of Deep Convolutional Generative Adversarial Network (DCGAN), Least Squares Generative Adversarial Network (LSGAN), and BigGAN [3-6]. This often leads to an unstable training process due to over-parameterization tuning [7, 8]. Conversely, reducing the number of tunable parameters can alleviate training difficulty and improve stability, as demonstrated by lightweight GAN variants like TinyGAN and Progressive Growing Generative Adversarial Network (PGGAN or ProGAN) [9, 10]. However, this reduction often comes at the cost of losing critical feature information, resulting in lower image quality. An alternative solution is to modify training strategies. For instance, models such as Self-Attention Generative Adversarial Network (SAGAN) and StyleGAN employ the Two Time-Scale Update Rule (TTUR), which carefully adjusts the learning rates and update frequencies for the generator and discriminator [11, 12].

Nonetheless, these tuning methods considerably increase the complexity of tunable hyperparameters and the overall training process, potentially leading to instability [13]. Although smoothing labels can be used to improve stability by reducing over-confidence of the discriminator, as employed in the one-sided and two-sided label smoothing [14, 15].

This paper, instead, proposes a new GAN model that leverages a convolutional neural network (CNN)-based architecture combined with the spectral normalization technique, to achieve a balance between training stability and image generation quality. A carefully designed normalization strategy is integrated into both the generator and discriminator parts of the Inception module of the CNN architecture, to stabilize the adversarial training process while maintaining the expressive capacity required for high-fidelity image synthesis. Collectively, we termed this Inception-inspired model the Inception Generative Adversarial Network (IGAN).

## 2. Related Work

The work in [3] proposed a category of generative adversarial networks (GANs) known as DCGAN. These networks are designed with CNN architecture and are very effective for unsupervised learning. Average pooling and batch normalization (BN) are used to help stabilize the learning process. Another large-scale GAN, known as BigGAN, which utilizes a different residual block structure, is proposed in [16]. Orthogonal regularization, self-attention mechanism, and spectral normalization are applied to stabilize training and improve the quality of generated images. However, the model is designed for large-scale datasets, and its performance may degrade when applied to smaller or less diverse datasets. Additionally, training BigGAN requires fine-tuning of hyperparameters, e.g., batch size, learning rate, and regularization parameters, which makes the training process more complex and less robust. A fully connected convolutional GAN (FCC-GAN) architecture is proposed in [17]. The generator employs deep fully connected (FC) layers to transform low-dimensional input noise into a high-dimensional intermediate representation of image features. It also utilizes convolution layers to convert these features into an output image. A discriminator consists of FC layers to transform high-dimensional data into a lower-dimensional space for classification. While FC layers aim to transform noise into a high-dimensional representation, this approach may lack the spatial hierarchy typically captured by convolutional layers alone, resulting in less realistic outputs. Also, the FCC-GAN architecture does not address mode collapse. The work in [18] introduced an alternative generator architecture for StyleGAN that contains a fully connected multi-layer perceptron (FC-MLP). StyleGAN used adaptive instance normalization (AdaIN) to modify the activations of the generator. By altering the styles according to size, the generator architecture enables control over the image synthesis. Although StyleGAN introduces improvements in training stability compared to earlier GANs, challenges such as convergence issues and mode collapse persist in certain situations.

Improved DCGAN architecture is reflected in two layers [19]: first, an upsampling layer in the generator output layer is used to increase the data size. Then, a dropout layer is included in each layer of the discriminator to address the issue of gradient vanishing in GANs. The modified DCGAN is tested on the MNIST dataset [20]. Another structure modification proposed a stable U-Net GAN (SUGAN) [21]. The generator is composed of residual blocks, each block utilizes a spectral and batch normalization layer. The discriminator utilized a U-Net structure, consisting of an encoder (downsampling) and a decoder (upsampling) with a skip connection between the encoder and decoder.

## 3. Methods

Motivated by the limitations of the related work discussed in Section 2, our proposed IGAN model is designed with deeper convolutional layers using untraditional convolutions, such as pointwise and dilation convolutions [22]. The generator and discriminator networks are designed to increase their depth and width, while minimizing computational cost when reducing the number of weights due to the use of 1×1 convolution and dilated convolution. This is achieved to align with the multi-scale processing principle, which is also known as the Application Programming Interface (API). The functional API is capable of managing models with non-linear topology, but it is a sparse structure that involves parallel execution of layers throughout modules instead of a sequential model. This way, our proposed architecture of the IGAN model consists of two modules in the generator and two modules in the discriminator. The structure of this IGAN model is inspired by the conventional Inception network module, but with three main differences: (i) in the input dimensions of features, (ii) the average pooling replaced the max pooling layers, and (iii) the dilation convolution is used with Conv 5×5 kernels (or filters). Figure 1 illustrates the sparse structure modules of the generator.

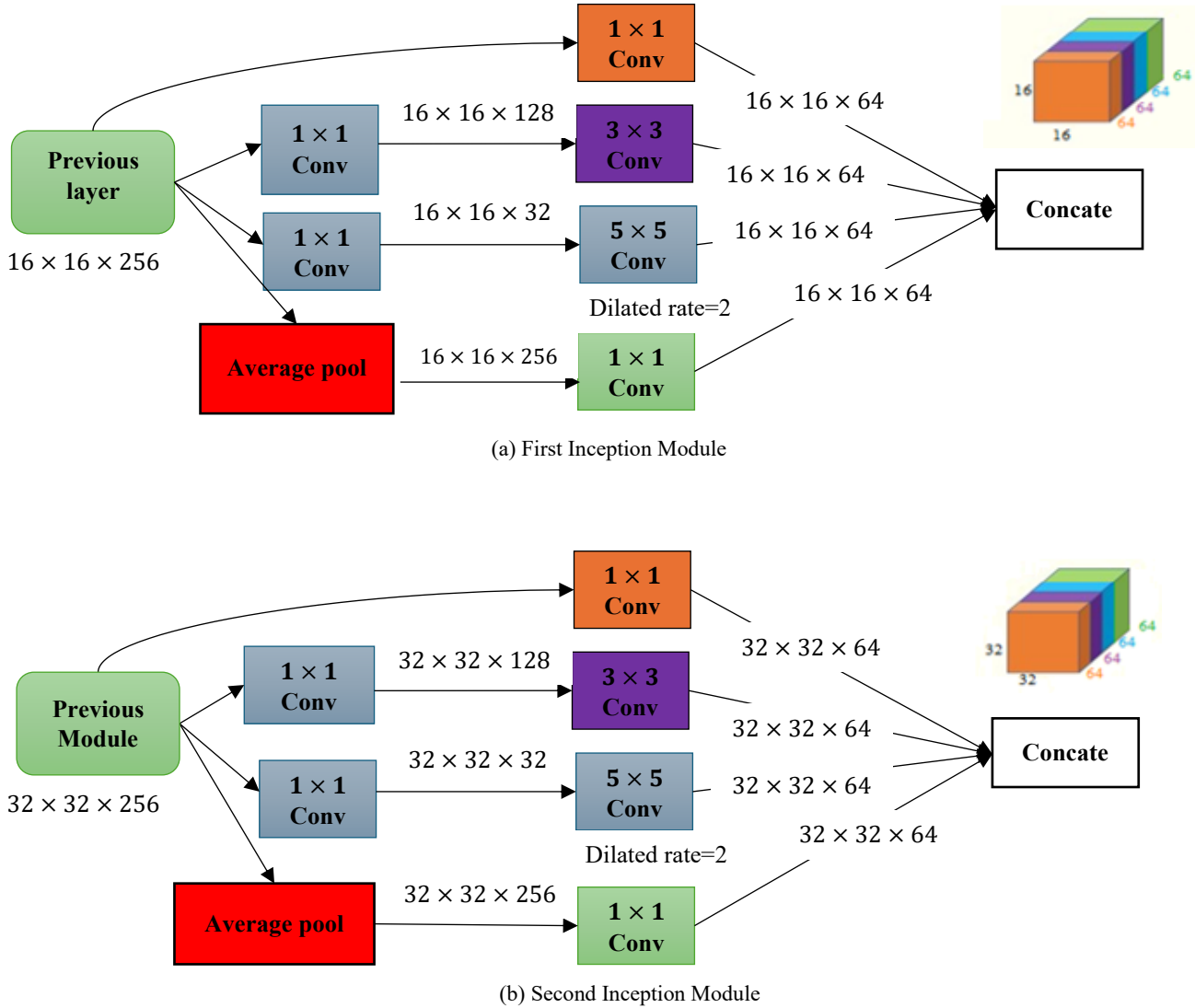

(a) First Inception Module

(b) Second Inception Module

**Figure 1.** The architecture of the sparse structure modules of the generator.

Each module is a combination of four parallel layers that only handle 1×1, 3×3, and 5×5 filter sizes; the input features from the previous convolution layer enter each layer inside the module, and their output is concatenated by the concatenate layer into a single output vector, forming the input to the next convolution layer or next inception block. The generator uses the upsampling technique to increase size dimensions until the image is 64×64. In discriminator, the dropout technique is used to overcome any overfitting. In Figure 1 using a combination of 1×1, 3×3, and 5×5 convolutions, as well as average pooling followed by concatenation aims to capture information at different scales and maintain computational efficiency:

1) 1×1 convolution reduces the number of channels and mixes channel-wise features without affecting the spatial resolution that helps control computational cost.

2) 3×3 convolution is a sweet spot for capturing local textures and spatial details, crucial for generating high-quality images in the generator.
3) 5×5 convolution (larger kernel sizes) helps the GAN generator understand more extensive features, such as large shapes or objects or color gradients, which are essential for producing coherent images.
4) Average pooling can aggregate features spatially, improving robustness to small variations in the input image.

Concatenating features from different convolution layers allows GANs to merge information across scales, improving the ability to generate diverse textures and coherent global structures. Combining fine-grained features produces more realistic and high-fidelity outputs. By capturing multi-scale features, the discriminator can evaluate images in a richer, more nuanced manner. This ensures the generator is challenged to improve across all convolution levels, leading to more stable and robust adversarial training. Table 1 shows the architecture of the IGAN.

**Table 1.** The architecture of the IGAN for generating 64×64 images.

| Generator | Kernel size | Stride | Dilation rate | Parameters | Output size |
|---|---|---|---|---|---|
| Input Layer | - | - | - | 0 | (100) |
| Dense | - | - | - | 413696 | (4096) |
| Reshape | - | - | - | 0 | (4, 4, 256) |
| UpSampling2D | - | - | - | 0 | (8, 8, 256) |
| Conv2D + BN + ReLU+ SN | 3 ×3 | 1 | 2 | 590080 + 1024 | (8, 8, 256) |
| UpSampling2D | - | - | - | 0 | (16, 16, 256) |
| **First Inception module** + BN+ ReLU | 1×1 | 1 | 1 | 199,072 + 1024 | (16, 16, 256) |
|  | 3 ×3 | 1 | 1 |  |  |
|  | 5×5 | 1 | 2 |  |  |
| UpSampling2D | - | - | - | 0 | (32, 32, 256) |
| **Second Inception module** + BN+ ReLU | 1 ×1 | 1 | 1 | 199,072 + 1024 | (32, 32, 256) |
|  | 3 ×3 | 1 | 1 |  |  |
|  | 5×5 | 1 | 2 |  |  |
| UpSampling2D | - | - | - | 0 | (64, 64, 256) |
| Conv2D + BN + ReLU + SN | 3 × 3 | 1 | 1 | 295040 + 512 | (64, 64,128) |
| Conv2D + Tanh | 3 × 3 | 1 | 1 | 3459 | (64, 64,3) |
| **Discriminator** | **Kernel size** | **Stride** | **Dilation rate** | **Parameters** | **Output size** |
| Input Layer | - | - | - | - | (64, 64,3) |
| Conv2D + LeakyReLU + Dropout | 3×3 | 1 | 2 | 1792 | (64, 64, 64) |
| **First Inception module** + BN + Dropout+ LeakyReLU | 1×1 | 1 | 2 | 43200 + 512 | (32, 32, 128) |
|  | 3×3 | 2 | 1 |  |  |
|  | 5×5 | 2 | 1 |  |  |
| **Second Inception module** + BN + Dropout + LeakyReLU | 1×1 | 2 | 1 | 217536 + 1024 | (16, 16, 256) |
|  | 3×3 | 2 | 1 |  |  |
|  | 5×5 | 2 | 1 |  |  |
| Conv2D | 1 × 1 | 2 | 1 | 65792 | (8, 8, 256) |
| Conv2D + BN + Dropout + LeakyReLU | 3 × 3 | 2 | 1 | 590080+ 1024 | (4,4, 256) |
| Flatten | - | - | - | 0 | (4096) |
| Dense+ Sigmoid | - | - | - | 4097 | 1 |

Figure 2 and Figure 3 show the architecture and layers sequence of the generator and discriminator, respectively. IGAN was based on a novel, deeper architectural design, which is the use of parallel execution with the use of pointwise and dilation convolution as a block structure to make both GAN's networks deeper. Several techniques have been used with this new architecture together, such as the Spectral Normalization (SN) technique in some layers of the generator, batch normalization, dropout, and average pooling, as shown in Table 1.

Figure 2 shows the flow through the generator network. The layers start from noise vector (100), pass through FC to output (4096), reshape to (4×4×256), upsampling to (8×8×256), pass through 3×3 conv, upsampling to (16×16× 256), then pass to the first inception module that contains four parallel convolution levels: first is 1×1 conv, second is (1×1conv then → 3×3 conv), third (1×1conv then → 5×5), fourth (average pooling → 1×1). Then concatenate all convolutions to result in the output of the first module (16×16×256). The output is upsampled to (32×32×256), then passed to the second inception module, which is the same first module level, but the output is (32×32×256). Then the output is upsampled to (64×64×256), passed through conv to produce (64x64x 128), and conv to produce (64×64× 3). When 64×64 acts as the dimensions of the generated image, 3 is depth, which acts as the RGB color. Algorithm 1 illustrates the pseudocode of IGAN.

Figure 3 shows the flow through the discriminator network, the layers start from (64×64×3) that act as color image (RGB) with dimension 64×64, then pass through convolution to be (64×64 ×64), this feature enters the first inception module that contains four parallel convolution levels: first is 1x1 conv, second is (1x1conv then → 3×3 conv), third (1×1conv then → 5×5), fourth (average pooling → 1×1).

Then concatenate all convolutions to result in the output of the first module (32×32×128), then pass to the second module, which is the same first module level, but the output is (16×16×256) then pass through convolutions → (8×8×256) → (4×4×256) → then use fully connected (flatten) to be output as vector (4096) then pass fully connected to output 1 value that is been 0 or 1. Between layers, yet between modules dropout is used to overcome any overfitting that may occur.

## 4. Results

The proposed IGAN is trained on CUB 200 [23] and a total of 20 distinct categories of ImageNet [24], which are aggregated and subsequently trained individually on eight categories of Tiger, Saint Bernard, Geyser, Redshank, Bee, Ant, Valley, and Goldfish; the obtained generated images are highly realistic, as shown in Figure 4.

The loss functions graph of the generator and discriminator networks across 500 epochs is depicted in Figure 5, which consists of two graphs: (i) the application of the proposed IGAN algorithm on twenty categories of ImageNet, which consist of 10,336 images, and (ii) the application of the proposed IGAN algorithm on the CUB 200 dataset. Both loss functions show highly notable levels of fluctuations at the beginning of training. The discriminator's loss stabilizes rapidly, indicating its increasing proficiency in differentiating between genuine and counterfeit data. In contrast, the loss function of the generator exhibits more significant fluctuations, which steadily reduced afterwards, suggesting that the generator is becoming more skilled at generating authentic data that may successfully deceive the discriminator.

The overall trend illustrates the adversarial training dynamics, wherein both networks consistently adjust and enhance their performance in reaction to one another. We note that Figure 5(b) reflects a more stable and effective way to train the generator due to the existence of 10,000 images in only one class. To strengthen the generated image quality, training is performed over a total of 500 iterations.

The IGAN was compared with state-of-the-art GAN models: BigGAN, SNGAN, and SAGAN, since these algorithms are commonly applied to the image generation domain, which aligns with the typical data type used in the proposed IGAN algorithm, allowing for a fair comparison [9, 25, 26] as reported in Table 2.

All FID and IS values of the other benchmark models are obtained from corresponding reference papers. However, relevant Peak Signal-to-Noise Ratio (PSNR) and Structural Similarity Index (SSIM) values are computed by comparing the images generated by the baseline models in the indicated reference papers with real images generated by our model for the same reference image using our code to calculate PSNR and SSIM metrics. One can see that the IGAN obtained the best values in all metrics.

Table 2. Comparison of the IGAN with state-of-the-art GAN models.

| Algorithm | CUB 200 | | | | ImageNet | | | |
|---|---|---|---|---|---|---|---|---|
| | IS | FID | PSNR | SSIM | IS | FID | PSNR | SSIM |
| BigGAN | 4.98 | 18.30 | 9.24 | +0.867 | 38.05 | 22.77 | 8.49 | +0.914 |
| SNGAN | 5.41 | 47.75 | 9.00 | +0.702 | 32.25 | 26.79 | 9.38 | +0.913 |
| SAGAN | 5.48 | 54.29 | 7.28 | +0.882 | 29.85 | 34.73 | 9.47 | +0.855 |
| IGAN | 9.27 | 13.12 | 11.71 | +0.817 | 68.25 | 15.08 | 9.41 | +0.956 |

**Figure 2.** Architecture process diagram of the generator of the IGAN.

**Figure 3:** Architecture process diagram of the discriminator of the IGAN.

**Algorithm 1**. The pseudocode of the IGAN.

**Input: noise vector, batch of real images.**

**Return: G, D: Trained generator and discriminator.**

1. Initialize the Generator and Discriminator with random weights.

2. Repeat for multiple iterations.

    **A. Train the Discriminator**

  a) Select a batch of real images from the dataset and combine them with the fake images generated by the Generator.

  b) Discriminator consists of:

      1. Conv2D + LeakyReLU + Dropout

      2. Inception Modules ($1 \times 1$, $3 \times 3$, $5 \times 5$): Learn multi-scale features.

      3. Conv2D and Flatten layers.

      4. Dense Layer + Sigmoid: (real or fake).

  c) Calculate Discriminator loss

  d) Update the discriminator by backpropagation to update the discriminator's weights.

    **B. Train the Generator**

  a) Generate fake data: Pass random noise through the Generator.

  b) The Generator consists of:

      **1)** Input Layer: Noise vector (100).

      **2)** Dense Layer: Converts noise to a (4096) vector.

      **3)** Reshape: Converts to a (4, 4, 256) volume.

      **4)** UpSampling2D layers: Increase the resolution from (4, 4, 256) → (8, 8, 256) → (16, 16, 256) → (32, 32, 256) → (64, 64, 256).

      **5)** Conv2D + BN + ReLU + SN: (64, 64, 128)

      **6)** Inception Modules ($1 \times 1$, $3 \times 3$, $5 \times 5$): learn multi-scale features.

      **7)** Conv2D + Tanh: Final output layer to generate an image of shape (64, 64, 3) (RGB image)

  c) Calculate Generator loss

  d) Update the Generator by backpropagation to update the Generator's weights.

3. End loops of iterations.

4. Return generated images.

5. End.

Figure 6 depicts isolation SSIM values in a separate figure for clearer presentation. The values are normalized between 0 and 1, so that the indicated values are not covered by the wide numerical ranges of IS, FID, or PSNR.

## 5. Conclusions

To overcome several limitations of current Generative Adversarial Network (GAN) variations, such as mode collapse and training instability, this paper proposes a novel GAN model that combines inception-inspired and dilated convolutions, which is termed the Inception GAN (IGAN). This IGAN model generates high-quality synthetic images and maintains training stability. The structure of this IGAN model is inspired by the Inception module of conventional CNN architecture, with the inclusion of parallel execution along with pointwise and dilation convolution.

Numerically, the IGAN model achieves the Fréchet Inception Distance (FID) of 13.12 on the CUB-200 dataset and 15.08 on the ImageNet dataset, as a 28–33% FID improvement as compared to the state-of-the-art GAN variants, with the Inception Score (IS) of 9.27 and 68.25, respectively. In conclusion, this reflects an improved image diversity and generation quality for the proposed IGAN model, while providing a scalable and computationally-efficient framework.

| | | | | | | |
|---|---|---|---|---|---|---|
| CUB200<br>FID:15.4 | 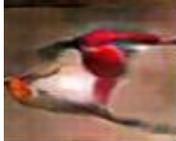 | 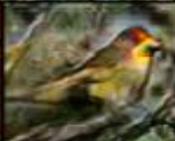 | 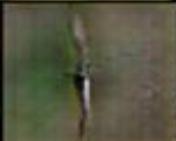 | 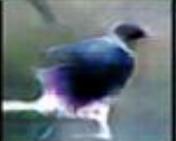 | 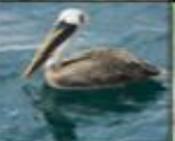 | 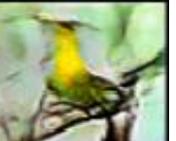 |
| Tiger<br>FID: 44.2 | 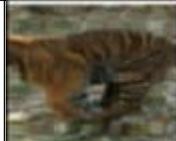 | 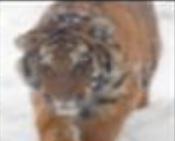 | 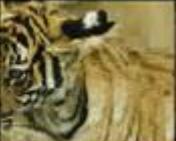 | 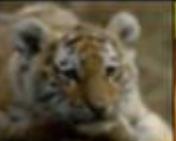 | 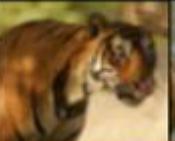 | 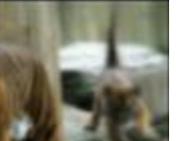 |
| Saint Bernard<br>FID: 34.7 | 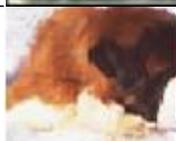 | 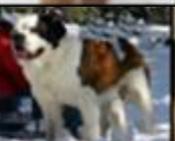 | 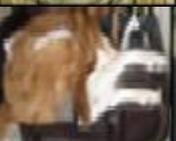 | 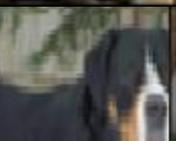 | 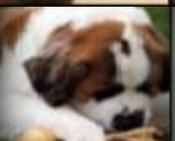 | 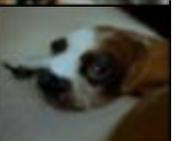 |
| Geyser<br>FID: 19.1 | 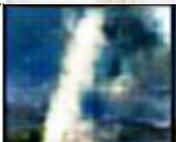 | 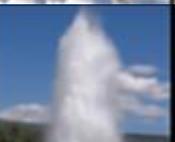 | 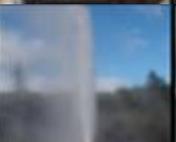 | 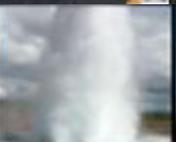 | 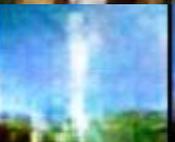 | 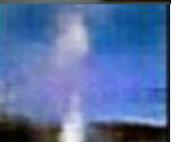 |
| Redshank<br>FID: 28.9 | 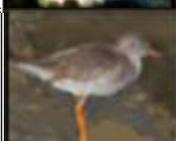 | 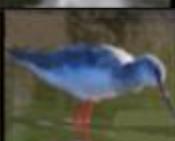 | 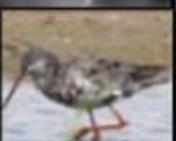 | 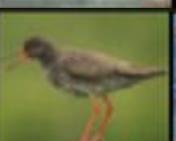 | 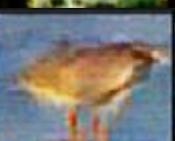 | 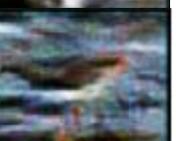 |
| Bee<br>FID: 29.3 | 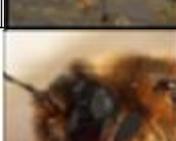 | 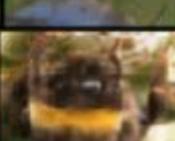 | 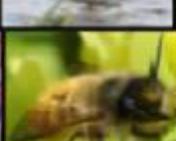 | 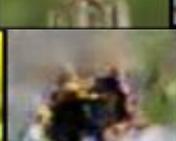 | 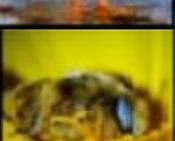 | 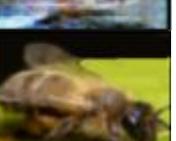 |
| Ant<br>FID: 40.2 | 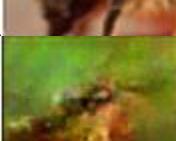 | 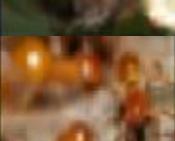 | 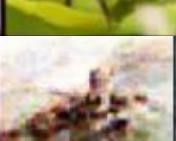 | 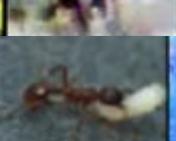 | 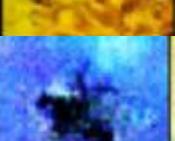 | 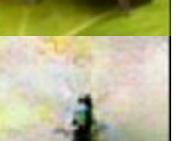 |
| Valley<br>FID: 24.4 | 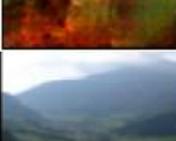 | 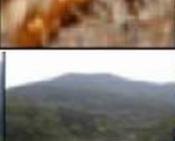 | 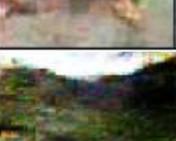 | 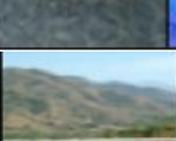 | 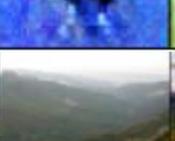 | 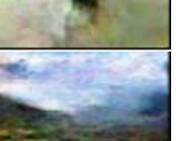 |
| Goldfish<br>FID: 17.4 | 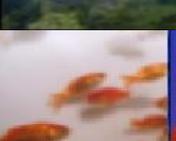 | 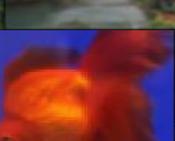 | 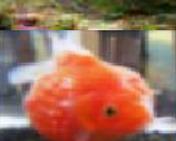 | 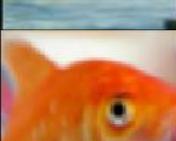 | 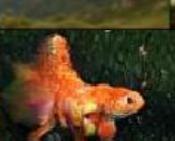 | 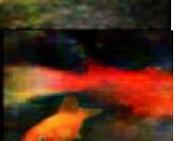 |

**Figure 4.** Example images generated by the IGAN with FID.

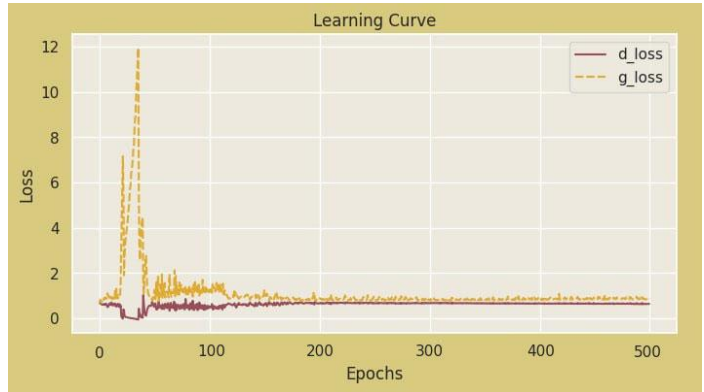

(a) ImageNet

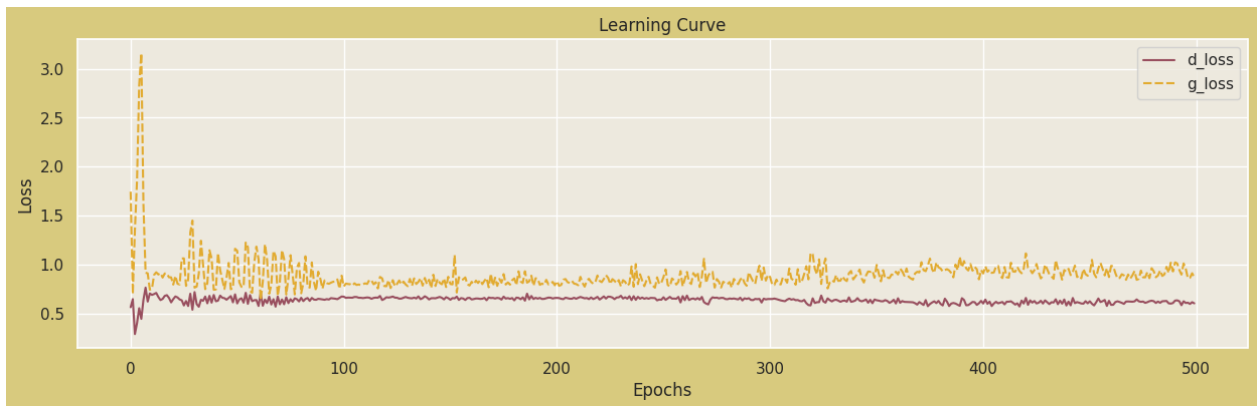

(b) CUB200

**Figure 5.** Loss function curves over the 500 Epochs using IGAN for the discriminator and generator on the CUB200 and twenty classes aggregated of ImageNet datasets.

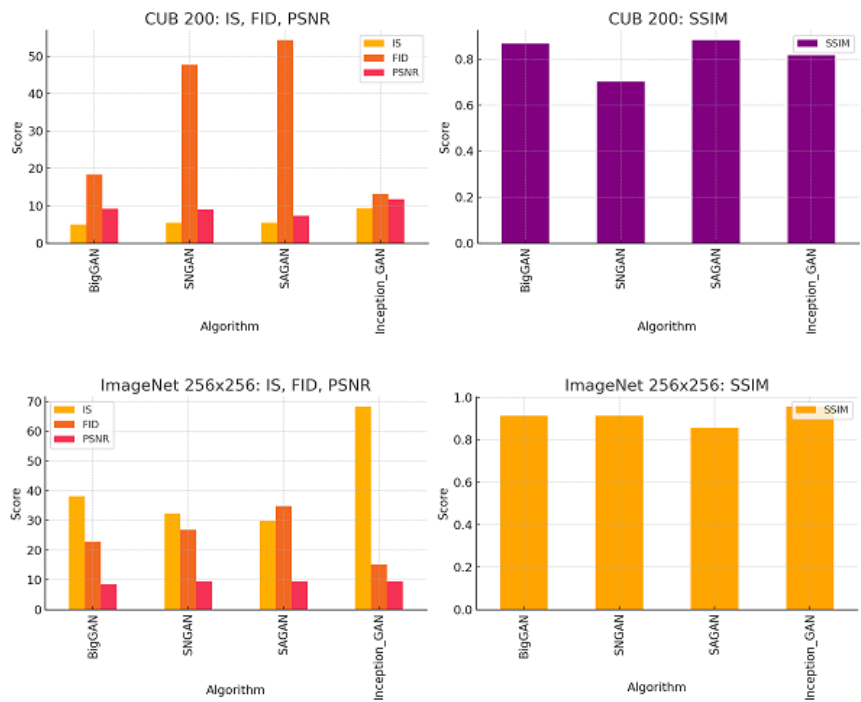

**Figure 6.** Algorithm comparison of SSIM, IS, FID, and PSNR across CUB 200 and ImageNet datasets.